# Comparison of the COG Defuzzification Technique and Its Variations to the GPA Index

Michael Gr. Voskoglou

Department of Mathematical Sciences, School of Technological Applications
Graduate Technological Educational Institute (T. E. I.) of Western Greece
Patras, Greece

**Abstract**

The Center of Gravity (COG) method is one of the most popular defuzzification techniques of fuzzy mathematics. In earlier works the COG technique was properly adapted to be used as an assessment model (RFAM) and several variations of it (GRFAM, TFAM and TpFAM) were also constructed for the same purpose. In this paper the outcomes of all these models are compared to the corresponding outcomes of a traditional assessment method of the bi-valued logic, the Grade Point Average (GPA) Index. Examples are also presented illustrating our results.

**Keywords:** *Grade Point Average (GPA) Index, Center of Gravity (COG) Defuzzification Technique. Rectangular Fuzzy Assessment Model (RFAM), Generalized RFAM (GRFAM), Triangular (TFAM) and Trapezoidal (TpFAM) Fuzzy Assessment Models.*

**1. Introduction**

*Fuzzy Logic (FL)*, due to its nature of characterizing the ambiguous situations of our day to day life by multiple values, offers rich resources for the assessment of such kind of situations. A characteristic example is the process of learning a subject matter, where the new knowledge is frequently connected to a degree of vagueness and/or uncertainty from the learner's, as well as the teacher's point of view.
In 1999 Voskoglou [10] developed a fuzzy model for the description of the process of learning a subject matter in the classroom in terms of the *possibilities* of the student profiles and later he assessed the student learning skills by calculating the corresponding system's *total possibilistic uncertainty* [11]. Meanwhile, Subbotin et al. [1], based on Voskoglou's model [10], adapted properly the frequently used in fuzzy mathematics *Center of Gravity (COG) defuzzification technique* and used it as an alternative assessment method of student learning skills. Since then, Voskoglou and Subbotin, working either jointly or independently, applied the COG technique and a number of variations of it for assessing several human or machine (Decision – Making, Case-Based Reasoning, etc.) skills, e.g. see [2-7, 12-16], , etc.
In the present paper the outcomes of the COG technique and its variations are compared to the corresponding outcomes of a traditional assessment method of the bi-valued logic, the *Grade Point Average (GPA)* index.
The rest of the paper is formulated as follows: In Section 2 we describe the classical GPA assessment method. In Section 3 we sketch the use of the COG technique as an assessment method, while in Section 4 we briefly describe the variations of the COG technique constructed in earlier papers and the reasons who led to the development of these variations. In Section 5 the outcomes of the COG technique and its variations are compared to the outcomes of the GPA index and examples are presented to illustrate our results. The last Section 6 is devoted to our conclusion and a discussion on the perspectives for future research on the subject.

**2. Traditional Assessment Methods**

The assessment methods which are commonly used in practice are based on principles of the bi-valued logic. The calculation of the mean value of the scores achieved by each one of its members is the classical method for assessing the *mean performance* of a group of objects (e.g. students, players, machines, etc.) with respect to an action. On the other hand, a very popular in the USA and other Western countries assessment method is the calculation of the *Grade Point Average (GPA)* index. This index is a weighted average in which greater coefficients (weights) are assigned to the higher scores. GPA, which is connected to the *quality group's performance*, is calculated by the formula GPA = $\frac{0 n_F + 1 n_D + 2 n_C + 3 n_B + 4 n_A}{n}$ (1), where $n$ is the total number of the group's members and $n_A$, $n_B$, $n_C$, $n_D$ and $n_F$ denote the numbers of the group's members that demonstrated excellent (A), very good (B), good (C), fair (D) and unsatisfactory (F) performance respectively [8].

In case of the worst performance ($n_F = n$) formula (1) gives that GPA = 0, while in case of the ideal performance ($n_A = n$) it gives GPA = 4. Therefore we have in general that $0 \leq$ GPA $\leq 4$. Consequently, values of GPA greater than 2 indicate a more than satisfactory performance.

Finally note that formula (1) can be also written in the form

GPA = $y_2 + 2y_3 + 3y_4 + 4y_5$ (2), where $y_1 = \frac{n_F}{n}$, $y_2 = \frac{n_D}{n}$, $y_3 = \frac{n_C}{n}$, $y_4 = \frac{n_B}{n}$ and $y_5 = \frac{n_A}{n}$ denote the frequencies of the group's members which demonstrated unsatisfactory, fair, good, very good and excellent performance respectively.

**3. The COG Defuzzification Technique as an Assessment Method (RFAM)**

The solution of a problem in terms of FL involves in general the following steps:
- *Choice of the universal set U* of the discourse.
- *Fuzzification* of the problem's data by defining the proper membership functions.
- *Evaluation of the fuzzy data* by applying rules and principles of FL to obtain a unique fuzzy set, which determines the required solution.
- *Defuzzification* of the final outcomes in order to apply the solution found in terms of FL to the original, real world problem.

One of the most popular in fuzzy mathematics defuzzification methods is the *Centre of Gravity (COG)* technique. For applying this method, let us assume that
A = *{(x, m(x)): x∈ U}* is the final fuzzy set determining the problem's solution. We correspond to each $x \in U$ an interval of values from a prefixed numerical distribution, which actually means that we replace *U* with a set of real intervals. Then, we construct the graph of the membership function *y=m(x)* and we consider the level's area F contained between this graph and the OX axis. There is a commonly used in FL approach (e.g. see [9]) to represent the system's fuzzy data by the coordinates ($x_c$, $y_c$) of the COG, say $F_c$, of the area $F$, which we calculate using the following well-known [19] from Mechanics formulas:

$$x_c = \frac{\iint_F x\,dxdy}{\iint_F dxdy}, \quad y_c = \frac{\iint_F y\,dxdy}{\iint_F dxdy} \quad (3).$$

Consider now the special case where one deals with the assessment of a group's performance Then, we choose as set of the discourse the set $U$ = {A, B, C, D, F} of the *fuzzy linguistic labels* (characterizations) of excellent (A), very good (B), good (C), fair (D) and unsatisfactory (F) performance respectively of the group's members. When a score, say $y$, is assigned to a group's member (e.g. a mark in case of a student), then its performance is characterized by F, if y $\in$ [0, 1), by D, if y $\in$ [1, 2), by C, if y$\in$ [2, 3), by B if y $\in$ [3, 4) and by A if y $\in$ [4, 5] respectively. Consequently, we have that $y_1 = m(x) = m(F)$ for all $x$ in [0,1), $y_2 = m(x) = m(D)$ for all $x$ in [1,2), $y_3 = m(x) = m(C)$ for all $x$ in [2, 3), $y_4 = m(x) = m(B)$ for all $x$ in [3, 4) and $y_5 = m(x) = m(A)$ for all $x$ in [4,5]. Therefore, the graph of the membership function $y = m(x)$, takes the form of Figure 1, where the area of the level's section $F$ contained between the graph and the OX axis is equal to the sum of the areas of the rectangles $S_i$, i=1, 2, 3, 4, 5.

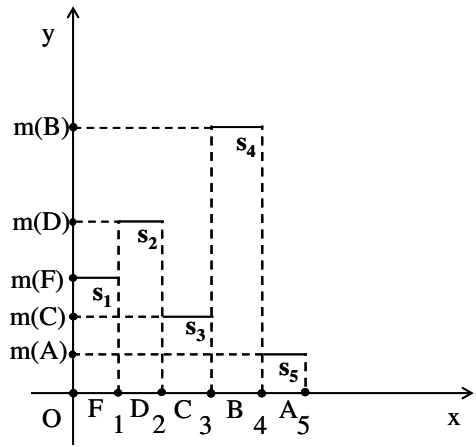

**Figure 1:** The graph of the COG method

It is straightforward then to check (e.g. see Section 3 of [12]) that in this case formulas (3) take the form:

$$x_c = \frac{1}{2}(y_1+3y_2+5y_3+7y_4+9y_5), \quad y_c = \frac{1}{2}(y_1^2+y_2^2+y_3^2+y_4^2+y_5^2) \quad (4),$$

with $x_1$=F, $x_2$=D, $x_3$=C, $x_4$=B, $x_5$=A and $y_i = \dfrac{m(x_i)}{\sum_{j=1}^{5} m(x_j)}$, i = 1, 2, 3, 4, 5. Note that the membership function $y = m(x)$, as it usually happens with fuzzy sets, can be defined, according to the user's choice, in any compatible to the common logic way. However, in order to obtain assessment results compatible to the corresponding results of the GPA index, we define here $y = m(x)$ in terms of the frequencies, as in formula (2) of Section 2. Then $\sum_{i=1}^{5} m(x_i) = 1$ (100%).

Using elementary algebraic inequalities and performing elementary geometric observations (e.g. Section 3 of [12]) one obtains the following assessment criterion:
- *Among two or more groups the group with the biggest $x_c$ performs better.*
- *If two or more groups have the same $x_c \geq 2.5$, then the group with the higher $y_c$ performs better.*
- *If two or more groups have the same $x_c < 2.5$, then the group with the lower $y_c$ performs better.*

As it becomes evident from the above statement, a group's performance depends mainly on the value of the x-coordinate of the COG of the corresponding level's area, which is calculated by the first of formulas (4). In this formula, greater coefficients (weights) are assigned to the higher grades. Therefore, the COG method focuses, similarly to the GPA index, on the group's *quality performance*. In case of the ideal performance ($y_5 = 1$ and $y_i = 0$ for $i \neq 5$) the first of formulas (4) gives that $x_c = \frac{9}{2}$. Therefore, values of $x_c$ greater than $\frac{9}{4} = 2.25$ demonstrate a more than satisfactory performance.

Due to the shape of the corresponding graph (Figure 1) the above method was named as the *Rectangular Fuzzy Assessment Model (RFAM)*.

## 4. Variations of the COG Technique (GRFAM, TFAM and TpFAM)

A group's performance is frequently represented by numerical scores in a climax from 0-100. These scores can be connected to the linguistic labels of *U* as follows: A (85-100), B(75-84), C (60-74), D(50-59) and F (0-49)[*].

Ambiguous cases appear in practice, being at the boundaries between two successive assessment grades; e.g. something like 84-85%, being at the boundaries between A and B. In an effort to treat better such kind of cases, Subbotin [4] "moved" the rectangles of Figure 1 to the left, so that to share common parts (see Figure 2). In this way, the ambiguous cases, being at the common rectangle parts, belong to both of the successive grades, which means that these parts must be considered *twice* in the corresponding calculations.

The graph of the resulting fuzzy set is now the bold line of Figure 2. However, the method used in Section 3 for calculating the coordinates of the COG of the area contained between the graph and the OX-axis is not the proper one here, because in this way the common rectangle parts are calculated only once. The right method for calculating the coordinates of the COG in this case was fully developed by Subbotin & Voskoglou [7] and the resulting framework was called the *Generalized Rectangular Fuzzy Assessment Model (GRFAM)*. The development of GRFAM involves the following steps:

1. Let $y_1, y_2, y_3, y_4, y_5$ be the *frequencies* a group's members who obtained the grades F, D, C, B, A respectively. Then $\sum_{i=1}^{5} y_i = 1$ (100%).

2. We take the heights of the rectangles in Figure 2 to have lengths equal to the corresponding frequencies. Also, without loss of generality we allow the sides of the

---

[*] This way of connection, although it satisfies the common sense, it is not unique; in a more strict assessment, for example, one could take A(90-100), B(80-89), C(70-79), D(60-69) and F (0-59), etc.

adjacent rectangles lying on the OX axis to share common parts with length equal to the 30% of their lengths, i.e. 0.3 units.[†]

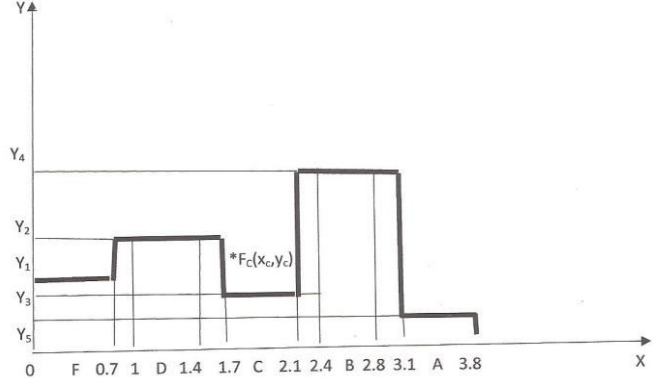

**Figure 2:** *Graphical representation of the GRFAM*

3. We calculate the coordinates ($x_{c_i}$, $y_{c_i}$) of the COG, say $F_i$, of each rectangle, i = 1, 2, 3, 4, 5 as follows: Since the COG of a rectangle is the point of the intersection of its diagonals, we have that $y_{c_i} = \frac{1}{2} y_i$. Also, since the x-coordinate of each COG $F_i$ is equal to the x- coordinate of the middle of the side of the corresponding rectangle lying on the OX axis, from Figure 2 it is easy to observe that $x_{c_i} = 0.7i - 0.2$.

4. We consider the system of the COGs $F_i$ and we calculate the coordinates ($X_c$, $Y_c$) of the COG F of the whole area considered in Figure 2 as the resultant of the system of the GOCs $F_i$ of the five rectangles from the following well known [20] formulas $X_c = \frac{1}{S}\sum_{i=1}^{5} S_i x_{c_i}$, $Y_c = \frac{1}{S}\sum_{i=1}^{5} S_i y_{c_i}$ (5).

In the above formulas $S_i$, i= 1, 2, 3, 4, 5 denote the areas of the corresponding rectangles, which are equal to $y_i$. Therefore $S = \sum_{i=1}^{5} S_i = \sum_{i=1}^{5} y_i = 1$ and formulas (5) give

that $X_c = \sum_{i=1}^{5} y_i (0.7i - 0.2)$, $Y_c = \sum_{i=1}^{5} y_i (\frac{1}{2} y_i)$ or

$X_c = (0.7\sum_{i=1}^{5} iy_i) - 0.2$, $Y_c = \frac{1}{2}\sum_{i=1}^{5} y_i^2$ (6).

5. We determine the area in which the COG F lies as follows: For i, j = 1, 2, 3, 4, 5, we have that $0 \leq (y_i - y_j)^2 = y_i^2 + y_j^2 - 2y_i y_j$, therefore $y_i^2 + y_j^2 \geq 2y_i y_j$, with the equality holding if, and only if, $y_i = y_j$. Therefore $1 = (\sum_{i=1}^{5} y_i)^2 =$

$= \sum_{i=1}^{5} y_i^2 + 2\sum_{\substack{i,j=1,\\ i \neq j}}^{5} y_i y_j \leq \sum_{i=1}^{5} y_i^2 + 2\sum_{\substack{i,j=1,\\ i \neq j}}^{5} (y_i^2 + y_j^2) = 5\sum_{i=1}^{5} y_i^2$ or $\sum_{i=1}^{5} y_i^2 \geq \frac{1}{5}$ (7), with the

equality holding if, and only if, $y_1 = y_2 = y_3 = y_4 = y_5 = \frac{1}{5}$. In case of the equality the first of formulas (6) gives that $X_c = 0.7(\frac{1}{5} + \frac{2}{5} + \frac{3}{5} + \frac{4}{5} + \frac{5}{5}) - 2 = 1.9$. Further,

---

[†] Since the ambiguous assessment cases are situated at the boundaries between the adjacent grades, it is logical to accept a percentage for the common lengths of less than 50%.

combining the inequality (7) with the second of formulas (6), one finds that $Y_c \geq \frac{1}{10}$

Therefore the *unique minimum for $Y_c$* corresponds to the COG $F_m$ (1.9, 0.1).

The *ideal case* is when $y_1 = y_2 = y_3 = y_4 = 0$ and $y_5=1$. Then formulas (2) give that $X_c = 3.3$ and $Y_c = \frac{1}{2}$. Therefore the COG in this case is the point $F_i$ (3.3, 0.5).

On the other hand, the *worst case* is when $y_1 = 1$ and $y_2 = y_3 = y_4 = y_5 = 0$. Then from formulas (2) we find that the COG is the point $F_w$ (0.5, 0.5). Therefore, the area in which the COG F lies is the area of the triangle $F_w\, F_m\, F_i$ (Figure 3).

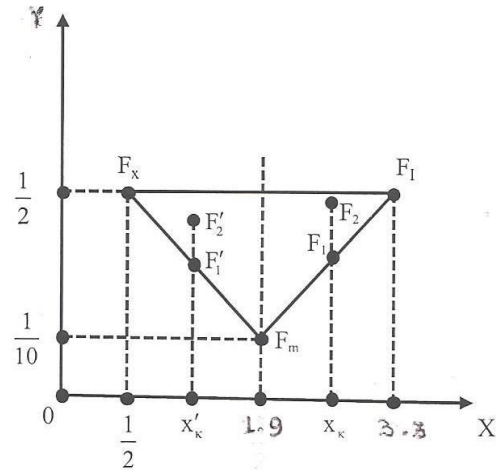

**Figure 3:** *The triangle where the COG lies*

6. From elementary geometric observations on Figure 3 one obtains the following assessment criterion:

- *Between two groups, the group with the greater $X_c$ performs better.*
- *If two groups have the same $X_c \geq 1.9$, then the group with the greater $Y_c$ performs better.*
- *If two groups have the same $X_c < 1.9$, then the group with the lower $Y_c$ performs better*

From the first of formulas (6) it becomes evident that the GRFAM measures the *quality* group's performance. Also, since the ideal performance corresponds to the value $X_c = 3.3$, values of $X_c$ greater than $\frac{3.3}{2} = 1.65$ indicate a more than satisfactory performance.

At this point one could raise the following question: Does the shape of the membership function's graph of the assessment model affect the assessment's conclusions? For example, what will happen if the rectangles of the GRFAM will be replaced by isosceles triangles? The effort to answer this question led to the construction of the *Triangular Fuzzy Assessment Model (TFAM)*, created by Subbotin & Bilotskii [2] and fully developed by Subbotin & Voskoglou [3].

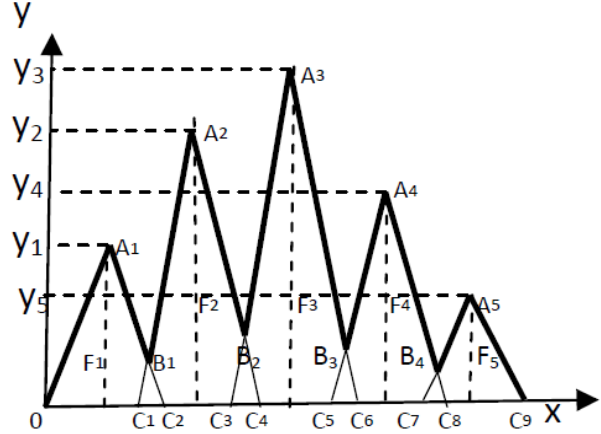

**Figure 4:** *Graphical Representation of the TFAM*

The graphical representation of TFAM is shown in Figure 4 and the steps followed for its development are the same with the corresponding steps of GRFAM presented above. The only difference is that one works with isosceles triangles instead of rectangles. The final formulas calculating the coordinates of the COG of TFAM are:

$$X_c = (0.7\sum_{i=1}^{5} iy_i) - 0.2, \qquad Y_c = \frac{1}{5}\sum_{i=1}^{5} y_i^2 \quad (8)$$

and the corresponding assessment criterion is the same with the criterion obtained for GRFAM.

An alternative to the TFAM approach is to consider isosceles trapezoids instead of triangles [4, 5]. In this case we called the resulting framework *Trapezoidal Fuzzy Assessment Model (TpFAM)*. The corresponding scheme is that shown in Figure 5.

In this case the $y$ - coordinate of the COG $F_i$, i=1, 2, 3, 4, 5, of each trapezoid is calculated in terms of the fact that the COG of a trapezoid lies on the line segment joining the midpoints of its parallel sides $a$ and $b$ at a distance $d$ from the longer side $b$ given by $d = \frac{h(2a+b)}{3(a+b)}$, where $h$ is its height [18]. Also, since the $x$-coordinate of the COG of each trapezoid is equal to the x-coordinate of the midpoint of its base, it is easy to observe from Figure 5 that $x = 0.7i - 0.2$.

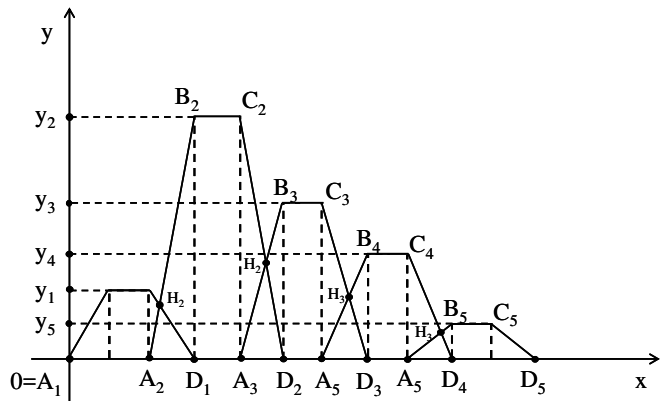

**Figure 5:** The TpFAM's scheme

One finally obtains from formulas (5) that $X_c = (0.7\sum_{i=1}^{5} iy_i) - 0.2$, $Y_c = \frac{3}{7}\sum_{i=1}^{5} y_i^2$ (9)

and the assessment criterion is the same again.

**5. Comparison of the Assessment Methods**

One can write formulas (6), (8) and (9) of Section 4 in the single form:

$$X_c = (0.7\sum_{i=1}^{5} iy_i) - 0.2, \quad Y_c = a\sum_{i=1}^{5} y_i^2 \qquad (10),$$

where $a = \frac{1}{2}$ for the GRFAM, $a = \frac{1}{5}$ for the TFAM and $a = \frac{3}{7}$ for the TpFAM.
Combining formulas (10) with the common assessment criterion stated in Section 4 one obtains the following result:

**5.1 THEOREM:** *The three variations of the COG technique, i.e. the GRFAM, the TFAM and the TpFAM are equivalent assessment models.*

Further, the first of formulas (10) can be written as

$X_c = 0.7(y_1 + 2y_2 + 3y_3 + 4y_4 + 5y_5) - 0.2 = 0.7 [(y_2 + 2y_3 + 3y_4 + 4y_5) + \sum_{i=1}^{5} y_i] - 0.2$.

Therefore, by formula (2) of Section 3, one finally gets that $X_c = 0.7(\text{GPA} + 1) - 0.2$, or $X_c = 0.7\text{GPA} + 0.5$ (11).

In the same way, the first of formulas (4) of Section 3 for RFAM can be written as

$x_c = \frac{1}{2}(y_1 + 3y_2 + 5y_3 + 7y_4 + 9y_5) = \frac{1}{2}(2\text{GPA} + 1)$, or $x_c = \text{GPA} + 0.5$ (12).

We are ready now to prove:

**5.2 THEOREM:** *If the values of the GPA index are different for two groups, then the GPA index, the RFAM and its variations (GRFAM, TFAM and TpFAM) provide the same assessment outcomes on comparing the performance of these groups.*

*Proof:* Let G and G´ be the values of the GPA index for the two groups and let $x_c$, $x_c´$ be the corresponding values of the x-coordinate of the COG for the RFAM. Assume without loss of generality that G>G´, i.e. that the first group performs better according to the GPA index. Then, equation (12) gives that $x_c > x_c´$, which, according to the first case of the assessment criterion of Section 3, shows that the first group performs also better according to the RFAM.

In the same way, from equation (11) and the first case of the assessment criterion of Section 4, one finds that the first group performs better too according to the equivalent assessment models GRFAM, TFAM and TpFAM.-

In case of the same GPA index we shall show the following result:

**5.3 THEOREM:** *If the GPA index is the same for two groups then the RFAM and its variations (GRFAM, TFAM and TpFAM) provide the same assessment outcomes on comparing the performance of these groups.*

*Proof:* Since the two groups possess the same value of the GPA index, equations (11) and (12) show that the values of $X_c$ and $x_c$ are also the same. Therefore, one of the last two cases of the assessment criteria of Sections 3 and 4 could happen. The possible values of x in these criteria lie in the intervals $[0, \frac{9}{2}]$ and $[0, 3.3]$ respectively, while the critical points correspond to the values $x_c = 2.5$ and $X_c = 1.9$ respectively. Obviously, if both values of $x$ are in $[0, 1.9)$, or in $[2.5, \frac{9}{2}]$, then the two criteria provide the same assessment outcomes on comparing the performance of the two groups. Assume therefore that $1.9 < X_c$ and $x_c < 2.5$. Then, due to equation (11), $1.9 < X_c \Leftrightarrow 1.9 < 0.7\text{GPA} + 0.5 \Leftrightarrow 1.4 < 0.7\text{GPA} \Leftrightarrow \text{GPA} > 2$.

Also, due to equation (12), $x_c < 2.5 \Leftrightarrow \text{GPA} + 0.5 < 2.5 \Leftrightarrow \text{GPA} > 2$. Therefore, the inequalities $1.9 < X_c$ and $x_c < 2.5$ cannot hold simultaneously and the result follows.-

Combining Theorems 5.2 and 5.3 one obtains the following corollary:

**5.4 COROLLARY:** *The RFAM and its variations GRFAM, TFAM and TpFAM provide always the same assessment results on comparing the performance of two groups.*

The following example shows that in case of the *same GPA values the application of the GPA index could not lead to logically based conclusions* (see also paragraph (vii) of Section 4 of [7]). Therefore, in such situations, our criteria of Sections 3 and 4 become useful due to their logical nature.

**5.5 EXAMPLE:** The student grades of two Classes with 60 students in each Class are presented in Table 1

**Table 1:** *Student Grades*

| Grades | Class I | Class II |
|---|---|---|
| C | 10 | 0 |
| B | 0 | 20 |
| A | 50 | 40 |

The GPA index for the two classes is equal to $\dfrac{2*10+4*50}{60} = \dfrac{3*20+4*40}{60} \approx 3.67$, which means that the two Classes demonstrate the same performance in terms of the GPA index. Therefore equation (11) gives that $X_c = 0.7*3.67 + 0.5 \approx 3.07$, while equation (12) gives that $x_c = 4.17$ for both Classes. But $\sum_{i=1}^{5} y_i^2 = (\dfrac{1}{6})^2 + (\dfrac{5}{6})^2 = \dfrac{26}{36}$ for the first and $\sum_{i=1}^{5} y_i^2 = (\dfrac{2}{6})^2 + (\dfrac{4}{6})^2 = \dfrac{20}{36}$ for the second Class. Therefore, according to the assessment criteria of Sections 3 and 4 the first Class demonstrates a better performance in terms of the RFAM and its variations.

Now which one of the above two conclusions is closer to the reality? For answering this question, let us consider the *quality of knowledge*, i.e. the ratio of the students received B or better to the total number of students, which is equal to $\dfrac{5}{6}$ for the first and 1 for the second Class. Therefore, from the common point of view, the situation in Class II is better. However, many educators could prefer the situation in Class I having a greater number of excellent students. Conclusively, in no case it is logical to accept that the two Classes demonstrated the same performance, as the calculation of the GPA index suggests.

The next example shows that although the RFAM, GRFAM, TFAM and TpFAM provide always the same assessment results on comparing the performance of two groups (Corollary 5.4), *they are not equivalent assessment models*.

**5.6 EXAMPLE:** Table 2 depicts the results of the final exams of the first term mathematical courses of two different Departments, say $D_1$ and $D_2$, of the School of Technological Applications (future engineers) of the Graduate T. E. I. of Western Greece. Note that the contents of the two courses and the instructor were the same for the two Departments.

**Table 2**: *Results of the two Departments*

| Grade | $D_1$ | $D_2$ |
|---|---|---|
| A | 1 | 1 |
| B | 3 | 6 |
| C | 11 | 13 |
| D | 9 | 10 |
| F | 6 | 5 |
| Total No. of students | 30 | 35 |

The GPA index is equal to $\frac{1*9+2*11+3*3+4*1}{30} \approx 1.47$ for $D_1$ and $\frac{1*10+2*13+3*6+4*1}{35} \approx 1.66$ for $D_2$. Therefore, the two Departments demonstrated a less than satisfactory performance (since GPA < 2), with the performance of $D_2$ being better.

Further, equation (11) gives that $X_c \approx 1.53$ for $D_1$ and $X_c \approx 1.66$ for $D_2$. Therefore, according to the first case of the assessment criterion of Section 4, $D_2$ demonstrated (with respect to GRFAM, TFAM and TpFAM) a better performance than $D_1$. Moreover, since $1.53 < \frac{3.3}{2} = 1.65 < 1.66$, $D_1$ demonstrated *a less than satisfactory performance*, while $D_2$ demonstrated *a more than satisfactory performance*.

In the same way equation (12) gives that $x_c \approx 1.97$ for $D_1$ and $x_c \approx 2.16$ for $D_2$. Therefore, according to the first case of the assessment criterion of Section 3, $D_2$ demonstrated (with respect to RFAM) a better performance than $D_1$. But in this case, since for both Departments $X_c < \frac{4.5}{2} = 2.25$, *the two Departments demonstrated a less than satisfactory performance*.

**REMARK:** Note that, if GPA > 2 (more than satisfactory performance), then
$X_c = 0.7\text{GPA} + 0.5 > 0.7 * 2 + 0.5 = 1.9 > 1.65$ and
$x_c = \text{GPA} + 0.5 > 0.2 + 0.5 = 2.5 > 2.25$. Therefore the corresponding group's performance is also more than satisfactory with respect to GRFAM, TFAM, TpFAM and RFAM.

However, if GPA < 2 (less than satisfactory performance), then $X_c < 1.9$ and $x_c < 2.5$, which do not guarantee that $X_c < 1.65$ and $x_c < 2.25$. Therefore the assessment characterizations of RFAM and the equivalent GRFAM, TFAM, TpFAM can be different only when GPA < 2.

## 6. Conclusion and Perspectives for Future Research

From the discussion performed in this paper it becomes evident that the RFAM and its variations GRFAM, TFAM and TpFAM, although they provide always the same assessment outcomes on comparing the performance of two groups, they are not equivalent assessment methods. The above assessment outcomes are also the same with those of the GPA index, unless if the groups under assessment possess the same values. In the last case the GPA index could not lead to logically based conclusions. Therefore, in this case either the use of RFAM or of its variations must be preferred.

Other fuzzy assessment methods have been also used in earlier author's works like the measurement of a system's uncertainty [11] and the application of the fuzzy numbers [17]. These methods, in contrast to the previous ones which focus on the corresponding group's quality performance, they measure its mean performance. The plans for our future research include the effort to compare all these methods in order to obtain the analogous conclusions.